*Article*

# Research on the Proximity Relationships of Psychosomatic Disease Knowledge Graph Modules Extracted by Large Language Models

**Zihan Zhou[1,5], Ziyi Zeng[1,5], Wenhao Jiang[2], Yihui Zhu[1,5,*], Jiaxin Mao[3], Yonggui Yuan[2], Min Xia[1,5], Shubin Zhao[4], Mengyu Yao[1,5], Yunqian Chen[1,5]**

1 School of Automation, Naning University of Information Science and Technology, Nanjing, 210044, China;
2 Department of Psychosomatics and Psychiatry, Zhongda Hospital, Medical School, Southeast University, Nanjing, Jiangsu, China, 210009
3 The First Affiliated Hospital of Ningbo University, Ningbo, Zhejiang, China, 315010
4 Nanjing LES Information Technology Co., Ltd. Nanjing, Jiangsu, China, 210014
5 Jiangsu Key Laboratory of Big Data Analysis Technology (B-DAT), Nanjing 210044. China
* Correspondence: zhuyh90@nuist.edu.cn

**Abstract:** As social changes accelerate, the incidence of psychosomatic disorders has significantly increased, becoming a major challenge in global health issues. This necessitates an innovative knowledge system and analytical methods to aid in diagnosis and treatment. Here, we establish the ontology model and entity types, using the BERT model and LoRA-tuned LLM for named entity recognition, constructing the knowledge graph with 9668 triples. Next, by analyzing the network distances between disease, symptom, and drug modules, it was found that closer network distances among diseases can predict greater similarities in their clinical manifestations, treatment approaches, and psychological mechanisms, and closer distances between symptoms indicate that they are more likely to co-occur. Lastly, by comparing the proximity d and proximity z score, it was shown that symptom-disease pairs in primary diagnostic relationships have a stronger association and are of higher referential value than those in diagnostic relationships. The research results revealed the potential connections between diseases, co-occurring symptoms, and similarities in treatment strategies, providing new perspectives for the diagnosis and treatment of psychosomatic disorders and valuable information for future mental health research and practice.

## 1. Introduction

In recent years, due to rapid economic development and a faster pace of life, the number of psychosomatic disorder patients has been increasing annually. Traditional treatment for psychosomatic disorders involves professional psychological counseling, yet most patients harbor fears and a resistant attitude towards it , and both patients and their relatives lack professional knowledge related to psychological healthcare, coupled with the concealment and complexity of psychosomatic disorders themselves, making it difficult for psychosomatic disorders to be promptly detected and intervened [1,2]. This necessitates an innovative knowledge system and analytical method, serving as a specialized data foundation for a medical information intelligent decision-making system, to assist in diagnosis and treatment [3,4].

The knowledge graph (KG) with a robust semantic network can establish a network framework for complex psychosomatic disorder data. The classic KG is a graph data structure composed of knowledge points [5-7]. Each knowledge point is represented by a subject-predicate-object triplet (SPO), where the subject (S) and object (O) describe entities, and the predicate (P) represents the relationship. The SPO triplet structure is suitable for expressing qualitative knowledge (for example, the triplet <S: olanzapine, P: treats, O: schizophrenia> can express that the drug Olanzapine can treat schizophrenia).

Psychiatric KG play an important role in various applications such as drug discovery, clinical decision support systems, and drug recommendations [8,9]. However, due to the lengthy text paragraphs, sparse data, and scattered knowledge in psychiatric datasets [10], there is scant research on KG in the field of psychiatry, and in the complex realm of psychosomatic disorders, solely relying on the construction of visual KG for information retrieval and intelligent questioning is inadequate. Consequently, present efforts fall short of fulfilling the requirements for psychiatric information management.

To fill this gap, this study employs the BERT model and large language models (LLM) for named entity recognition (NER) of psychiatric texts [11-17]. Then we constructs SPO triplets and creates a high-quality psychiatric KG. Next, we apply graph theoretical analysis methods to analyze the graph structure of the psychiatric KG. By focusing on the topological relationships between nodes related to diseases, symptoms, and drugs [18], we arrive at three conclusions: (1) Symptoms corresponding to diseases, diseases diagnosed through symptoms, and drugs required for diseases are clustered within localized modules [19,22,23]. (2) The similarity of diseases, co-occurrence of treatment strategies and symptoms are negatively correlated with their network distance [20]. (3) Modules featuring terminologized and non-terminologized symptoms (distinguished in this study through primary diagnosis/diagnosis, meaning diseases diagnosed primarily through professional terminologies and those diagnosed through non-professional terms) show that terminologized symptoms are more strongly associated with diseases based on their network proximity [21].

## 2. Materials and Methods

### *2.1. Dataset*

The field of psychosomatic disorders currently lacks standardized annotated data resources. Our data comes from 262 patient medical records in the psychiatric department of Zhongda Hospital in Nanjing, China. We performed data cleaning and preprocessing on the medical records and produced two differently formatted NER datasets. The first type of NER dataset uses the "BIO" (Begin: the beginning character of an entity, Inside: a middle or ending character of an entity, Outside: a character that is not part of an entity) scheme for annotation [24,25]. Each character in the text sequence is tagged with a "B-", "I-" and "O" label to indicate whether the character is part of a named entity. This dataset divides entities into 7 types. See Table 1.

**Table 1.** Entity types and their character representations

| Entity type | Label | Entity type | Label |
|---|---|---|---|
| patient | N | risk factor | R |
| symptom | Z | drug | M |
| severity | T | diagnosed disease | D |
| differential diagnosed disease | Y | | |

The second dataset is used for NER tasks with LLM. Initially, medical cases are divided into 16 sections based on different tasks, then dataset is constructed using each section separately. Examples of the dataset are shown in Figure 2.B. The 16 sections include Patient number, Chief complaint, Course of disease, Severity, Summary of primary symptoms, Admission status, Mental examination, Medical history, Risk factors, Inducement, Treatment plan, Primary diagnostic basis, Primary diagnosed disease, Differential diagnostic basis, Differential diagnosed disease and Summary of signs. An example of 16 sections is shown as follows:

(1) Patient number: 0101447813.

(2) Chief complaint: Unable to feel happy, gloomy and joyless. Unable to concentrate. Fatigued and weak, lack of energy. Having had negative thoughts. Irritable and anxious, feeling of panic and chest tightness. Trembling hands, headache. Irregular sleep patterns, day-night reversal. Poor appetite.

(3) Course of disease: A year ago, the patient began to experience low mood due to family issues and gastrointestinal disease, feeling persistently unhappy and unable to cheer up, with difficulty concentrating, unable to stop thinking about past sad events, feeling let down by parents, often self-blaming, feeling tired and weak, lacking energy, having passive thoughts but never acting on them, accompanied by symptoms of irritability, panic, chest tightness, trembling hands, and headaches. Irregular sleep patterns, reversal of day and night, poor appetite.

(4) Severity: Unable to live and study normally, took a leave of absence to recuperate the gastrointestinal tract at home, afterward the gastrointestinal disease gradually improved, but emotional problems worsened, preventing a normal return to study.

(5) Summary of primary symptoms: Symptoms include palpitations, chest tightness, trembling hands, and headaches. Insight is present. Always feeling down, unable to cheer up, difficulty concentrating, unable to control thoughts of past sad events, often self-blaming, feeling tired and weak, lacking energy, having negative thoughts but never acting on them, accompanied by symptoms of irritability, palpitations, chest tightness, trembling hands, and headaches. Sleep is irregular, with a reversed sleep cycle and poor appetite. Bowel and bladder functions are normal, sleep is average, and there has been no significant recent weight change.

(6) Admission status: Yesterday, visited the psychiatric outpatient clinic at Zhongda Hospital, where they diagnosed a "depressive episode" and admitted for treatment.

(7) Mental examination: Psychiatric evaluation: Conscious and oriented, passive engagement, rapid speech, no perceptual disturbances elicited, slightly active thinking, presence of pathological circumstantiality and obsessive thoughts, unstable emotions, decreased interest, poor concentration, impaired memory, reduced volition, poor appetite, sleep rhythm disturbances, with symptoms of palpitations, chest tightness, trembling hands, and headaches. Insight is present.

(8) Medical history: One year ago, the patient began to experience low mood due to family issues and gastrointestinal disease.

(9) Risk factors: Family factors: there are significant conflicts within the father's family, and the grandparents argue daily.

(10) Inducement: no obvious inducement.

(11) Treatment plan: The treatment plan should include antidepressants and mood stabilizers, supplemented with psychotherapy.

(12) Primary diagnostic basis: The patient speaks quickly because it is their natural manner, not because of the typical excitability associated with excessive talking.

(13) Primary diagnosed disease: Diagnosed with a depressive episode with mixed features.

(14) Differential diagnostic basis: The patient speaks quickly because it is their natural manner, not because of the typical excitability associated with excessive talking.

(15) Differential diagnosed disease: bipolar disorder, mania.

(16) Summary of signs: Sleep patterns are irregular, with day and night reversed, and appetite is poor. Emotional issues have gradually worsened. The patient is conscious and oriented, interacts passively, speaks quickly, exhibits no perceptual disturbances, has slightly active thoughts, pathologically circumstantial and obsessive thinking, unstable emotions, decreased interest, poor concentration, impaired memory, reduced volitional actions, poor appetite, disordered sleep rhythm, and symptoms including palpitations, chest tightness, trembling hands, and headaches. Insight is present.

The third dataset is used for querying text content. We first establish structured text content and then query the content of medical records. Ultimately, we hope that the LLM can use the content of the medical records or its own summarized statements to respond.

*2.2. Entity Recognition Technology Based on BERT*

The BERT language model [11,12], which utilizes the Transformer as a feature extractor [26-29], is a deeply trained bidirectional language model based on attention mechanisms. It has achieved outstanding results in various tasks in the Natural Language Processing (NLP) field, such as question answering systems, natural language inference, and NER, among others. This study uses the RoBERTa Chinese pre-trained character vector model, which is improved based on Google's official BERT pre-trained model for Chinese, and fine-tuned according to the NER task

Given a sentence $[W_1, W_2, \ldots, W_N]$, where $W_i$ represents a Chinese character in the sentence, and N represents the maximum length of the sentence. This sentence needs to be prefixed with a [CLS] token and suffixed with a [SEP] token before input, followed by character embedding through querying a character vector table, representing each character as a one-dimensional vector. For the NER task, the RoBERTa model is trained to predict the "BIO" labels for each character in the sequence. Therefore, we fine-tune the RoBERTa model with the first dataset, adding a classification layer to determine the label type of each character. The final pipeline is depicted in Figure 1.

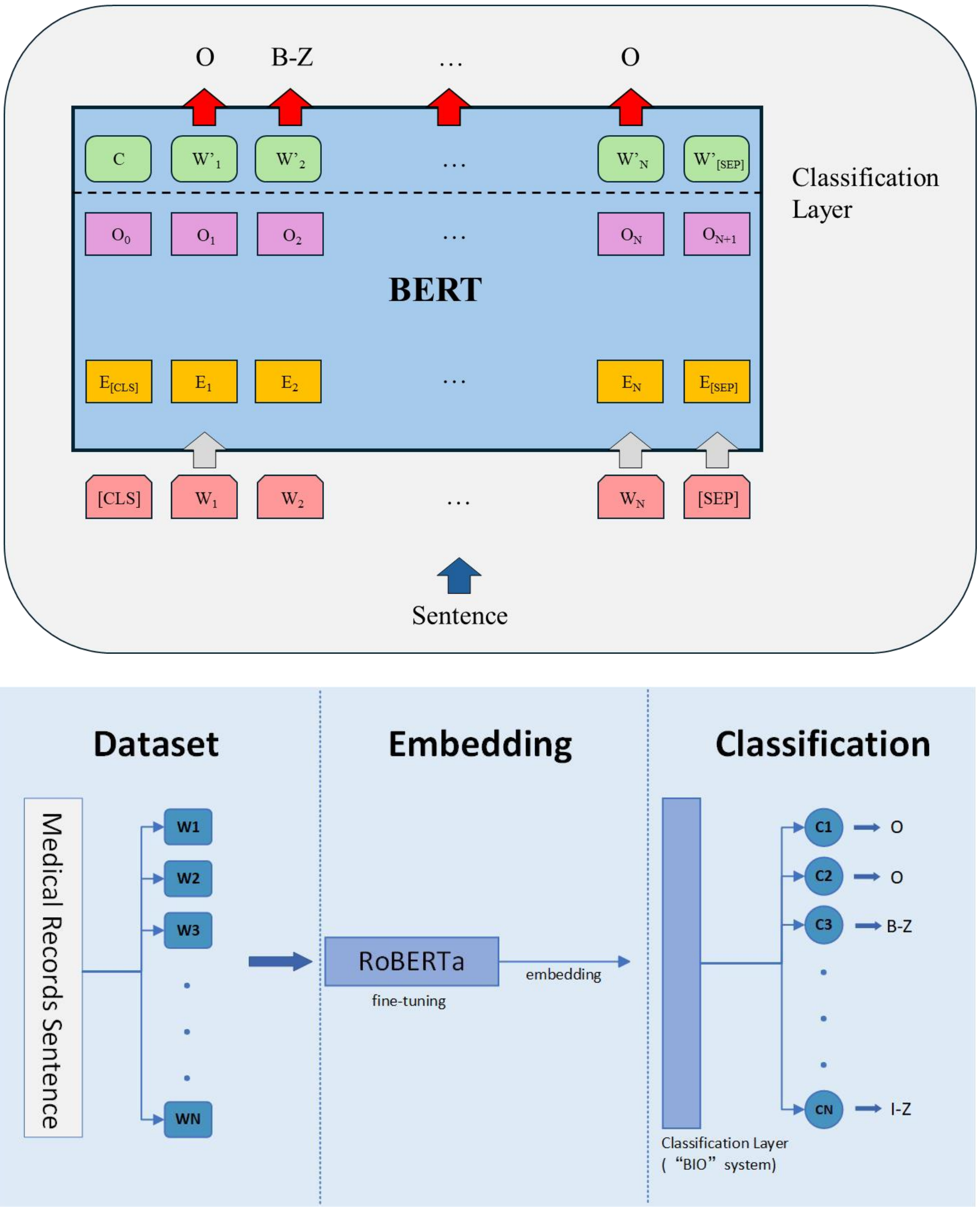

**Figure 1.** Fine-tuning the RoBERTa model for the NER task.

### *2.3. Entity Recognition Technology Based on Large Language Models*

LoRA is a low-rank adaptation fine-tuning method that significantly reduces the number of trainable parameters for downstream tasks and achieves excellent results [30]. For the pre-trained weight matrix $W_0 \in R^{d \times k}$ , it can be updated via low-rank decomposition $W_0 + \Delta W = W_0 + BA$ where $B \in R^{d \times r}$, $A \in R^{r \times k}$ and $r \ll min(d,k)$. During the training process, $W_0$ is frozen and does not receive gradient updates, while $A$ and $B$ contain trainable parameters. Figure 2.A is a schematic diagram of the LoRA principle, where matrix $A$ is initialized with random Gaussian values, and matrix $B$ is initialized to zero. When the input is $x$, for $h = W_0 x$, the modified forward propagation:

$$h = W_0 x + \Delta W x = W_0 x + BAx\#(1)$$

We use both the pre-trained LLaMA model and the LLaMA model fine-tuned with LoRA from the KnowLM [31-36] for entity extraction. This method uses the second type of dataset we created. When performing NER tasks with this project, it requires inputs of "Instruction" and "Input". "Instruction" provides detailed information about the task to be performed by the model. "Input" is the text information received by the LLM, forming the basis for the model to perform calculations and generate outputs. Subsequently, the LLM returns an "Output" result, which is determined by the "Instruction" and "Input" provided by the user. In this research, we define "Instruction" as "Extract possible entities and their types from the given text, with optional entity types are ['patient', 'symptom', 'drug', 'diagnosed disease', 'differential diagnosed disease', 'risk factor', 'severity'], answer in the format of (entity, entity type)". The final pipeline is depicted in in Figure 2.B.

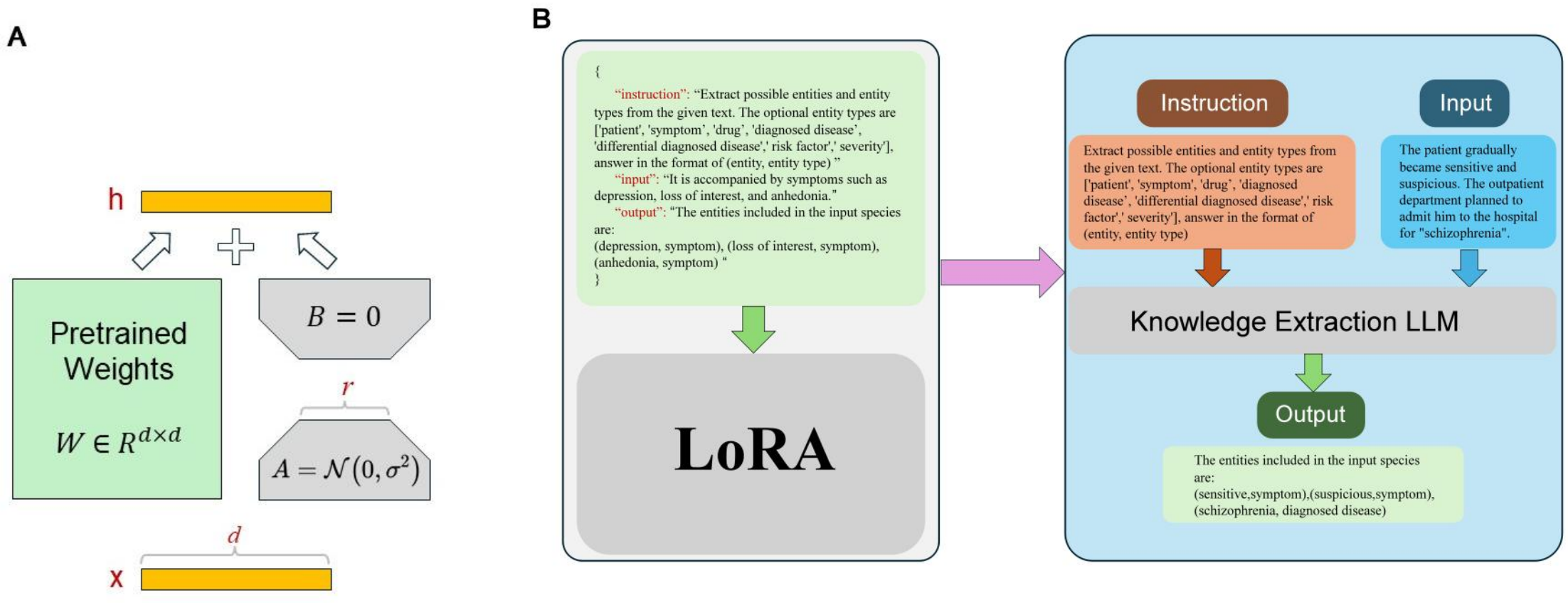

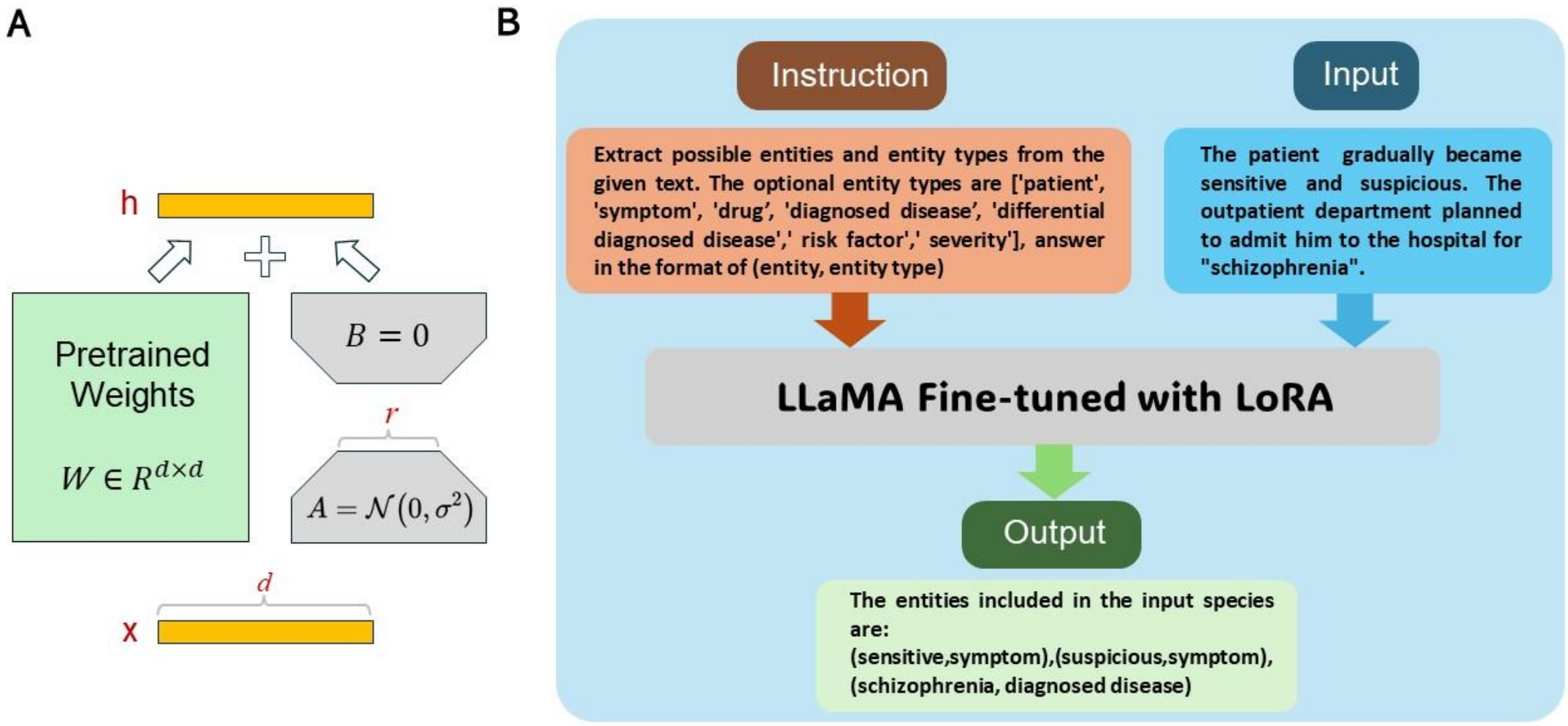

**Figure 2. (A)**Schematic diagram of the LoRA principle. **(B)** Process for LLM to extract entities and their types. First, the LLM is fine-tuned with LoRA using the second dataset, and upon completion of training, the fine-tuned LLM is used to perform the NER task.

### *2.4. Entity Alignment*

Through entity extraction, we obtained approximately 30,000 entities, which had a significant amount of duplication (for instance, if both case A and case B contain the entity "schizophrenia", then the entity "schizophrenia" would be extracted from both cases) and errors. Therefore, we performed data cleaning and integration on the extracted entities [37]. Then, we undertook a crucial data processing step, namely the standardization and unification of entity names [38-40]. This step was accomplished by calculating the degree of approximation between entities, including textual and semantic similarity of entities. Textual similarity of entities was determined by calculating the proportion of identical characters between two entities to decide if they represent the same expression [10]. The similarity was calculated using the Jaccard coefficient:

$$sim_t(A,B) = \frac{|A \cap B|}{|A \cup B|} \#(2)$$

Where $sim_t(A,B)$ represents the textual similarity between entity $A$ and entity $B$, $|A \cap B|$ represents the number of identical characters is between the two entities, and $|A \cup B|$ represents the total number of characters for both entities.

Semantic similarity calculates the similarity between two entities based on their approximation in contextual semantics. The method involves calculating the cosine of the word vectors between entities. The closer the cosine value is to 1, the smaller the vector angle distance, and the greater the approximation between the two entities:

$$sim_s(A,B) = cos\,\theta = \frac{\sum_i^n A_i \times B_i}{\sqrt{\sum_i^n (A_i)^2} \times \sqrt{\sum_i^n (B_i)^2}} \#(3)$$

Here, $sim_s(A,B)$ represents the semantic similarity between entity A and entity B, $A = [A_1, A_2, \dots, A_n]$ represents the word vector set of entity A, and $B = [B_1, B_2, \dots, B_n]$ represents the word vector set of entity B.

Due to the excessive similarity in the text of some different entities, relying solely on textual similarity can easily lead to errors in entity alignment. For example, "bipolar disorder" and "depression" are two different diseases, but they have a high textual similarity in Chinese. Therefore, we calculate entity similarity by assigning weights of 0.4 and 0.6 to textual similarity and semantic similarity, respectively:

$$sim(A,B) = sim_t(A,B) \times 0.4 + sim_s(A,B) \times 0.6\#(4)$$

For entities with high similarity, their entity names are standardized and unified. For example, the expression "bad temper" is used to replace similar terms like " lose temper easily" and "irritable". We merge the two entity types "diagnosed disease" and "differential diagnosed disease" into a single "disease" entity type. They will be differentiated by employing three distinct relationships: "primary diagnosis", "diagnosis", and "differential diagnosis". Ultimately, 3285 entities were obtained. Through such standardized processing, the KG can organize and link information more effectively. This improves the accuracy of queries and the search experience for users. It also enhances the quality of the graph, laying a solid foundation for subsequent analysis of the graph structure.

### *2.5. LCC and LCC z score*

The z-score of the Largest Connected Component (LCC) [41] of a set of nodes is used to describe the positioning of node sets in the psychosomatic disorder knowledge network. We calculate the size of the LCC formed by the set of nodes, and then compare the calculated LCC size with the expected LCC of a randomly selected set of nodes. The LCC z-score is the difference between the LCC size and the mean of randomization μ(random LCC), divided by the SD of the randomization σ(random LCC):

$$z_{LCC} = \frac{Observed\ LCC\ size - \mu(randomLCC)}{\sigma(randomLCC)} \#(5)$$

An LCC z-score greater than the expected mean indicates that the observed LCC is significantly larger than expected, meaning the node set aggregates into a local module.

*2.6. Network Distance $D_{ab}$ and Network Separation $S_{ab}$*

We measure the network relationship between two sets of nodes using network distance $D_{ab}$ and network separation $S_{ab}$ [42,43]. Network distance $D_{ab}$, also denoted as $<d_{ab}>$, is the average network distance between all pairs of nodes in the two sets of nodes:

$$D_{ab} = \frac{1}{|A| \times |B|} \sum_{a \in A, b \in B} d(a, b) \#(6)$$

Where $|A|$ and $|B|$ represents the number of nodes in two sets of nodes. Network separation compares the average shortest distance within each node set $<d_{aa}>$ and $<d_{bb}>$ with the average shortest distance $<d_{ab}>$ between node sets $A$ and $B$:

$$S_{ab} = < d_{ab} > - \frac{< d_{aa} > + < d_{bb} >}{2} \#(7)$$

The random expectation of $S_{ab}$ is zero; a negative $S_{ab}$ indicates that the two sets of nodes are in the same network neighborhood, while a positive $S_{ab}$ indicates that the two sets of nodes are topologically separated.

*2.7. Semantic Similarity*

We define semantic similarity to evaluate the biological and psychological mechanism similarities between diseases and between drugs [44,45]. we constructed an association matrix. Taking the symptom module of diagnosed diseases as an example, where each row represents a disease, each column represents a symptom, and the elements in the matrix (0 or 1) indicate whether there is an association between them. Using this association matrix, we calculate semantic similarity using the Wang method:

$$sim_{Wang}(g1, g2) = \frac{2 \times \sum_{t \in T_{g1} \cap T_{g2}} IC(t)}{IC(g1) + IC(g2)} \#(8)$$

Where $g1$ and $g2$ are two diseases, $T_{g1}$ and $T_{g2}$ represent the sets of symptoms or drugs corresponding to diseases $g1$ and $g2$, respectively, and $IC(g)$ denotes the number of symptoms or drugs associated with disease $g$.

*2.8. Network Proximity Distance and z score*

We define the network proximity metric (referred to in the text as "proximity distance d") as the average distance from all points in node set $A$ to the nearest points in node set B among two sets of nodes [42,43]:

$$d(A, B) = \frac{1}{|A|} \sum_{a \in A} min_{b \in B}\ dist(a, b) \#(9)$$

Where $|A|$ represents the number of nodes in node set $A$, $dist(a, b)$ represents the shortest distance between nodes a and b. Then, we simulated and obtained the expected distances between randomly selected disease-symptom pairs. We denote the expected mean distance as $\mu_{rand}(A, B)$, the SD as $\sigma_{rand}(A, B)$, and define the proximity z score:

$$z(A, B) = \frac{d(A, B) - \mu_{rand}(A, B)}{\sigma_{rand}(A, B)} \#(10)$$

The proximity z score measures the difference between the proximity distance and the expectation, with $z < 0$ indicating closer than random, and $z > 0$ indicating farther than random. For the proximity distance d and the proximity z score, lower metric values indicate closer distances between the two sets of nodes in the network. Since it's based on random simulation, the proximity z score is a stochastic measure, meaning the same repeated calculation can produce different proximity z score.

*2.9 Knowledge Base Construction*

This paper uses the LangChain framework to construct the knowledge base [46]. The workflow is shown in Figure 8. The workflow of LangChain is a highly integrated and automated process that includes six stages. (1) files need to be loaded from the local knowledge base, which is the first step in obtaining raw data. Then, the loaded file content is read and converted into a processable text format, and then the text is segmented according to certain rules (such as paragraphs, sentences, words). (2) Using NLP technology, the segmented text is converted into numerical vectors. Then, the converted text vectors are stored in a vector database. (3) Users' queries or questions also need to be converted into vectors using the same vectorization method to ensure comparison within the same vector space as the text vectors. (4) By calculating similarity measures, the Top k vectors most similar to the query vector are retrieved from the text vectors. (5) Using the retrieved text vectors, build context related to the question, and then enter them as a "Prompt" along with the question. (6) Finally, the question and context are submitted to LLM to generate an answer.

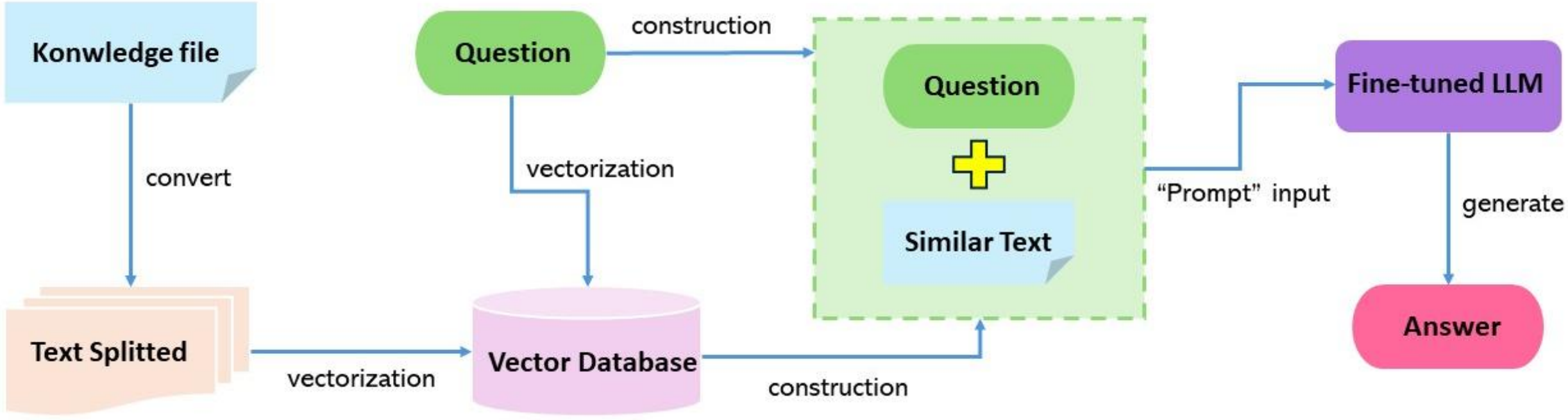

**Figure 8.** the workflow of Langchain

## 3. Results

*3.1. Construction of the Psychosomatic Disorder KG*

The KG on psychosomatic disorders comprises 3285 entities and 9668 triplets. The categories and quantities of entities are as follows: patient 189, symptom 2693, risk factor 75, severity 136, disease 94, drug 98. Entities in the patient category are represented by codes, such as "J+0101389333". There are 7 types of relationships: <patient, suffer, disease>, <drug, treat, disease>, <symptom, diagnosis, disease>, <symptom, primary diagnosis, disease>, <symptom, differential diagnosis, disease>, <risk factor, cause, disease>, <severity, diagnosis, disease>. See Figure 3.A.

The KG displays the relationships between diseases and related entities, visually characterizing the features of the psychosomatic disorder domain. In the graph, dots represent different entities, the color of the dots indicates the types of entities, and the lines between dots represent the relationships between entities. Taking "depression" as an example from the disease category. See Figure 3.B. The graph shows entities related to the disease type "depression" and their relationships, such as the <symptom: bad temper, diagnosis, disease: depression>, <drug: sertraline, treat, disease: depression>, <risk factor: high work pressure, cause, disease: depression>. See Table 2. Figure 3.C and Figure 3.D show examples of the structural network formed by the KG of symptom-disease pairs and drug-disease pairs, with subsequent graph structure analysis based on the structural network formed by the nodes of the KG.

A

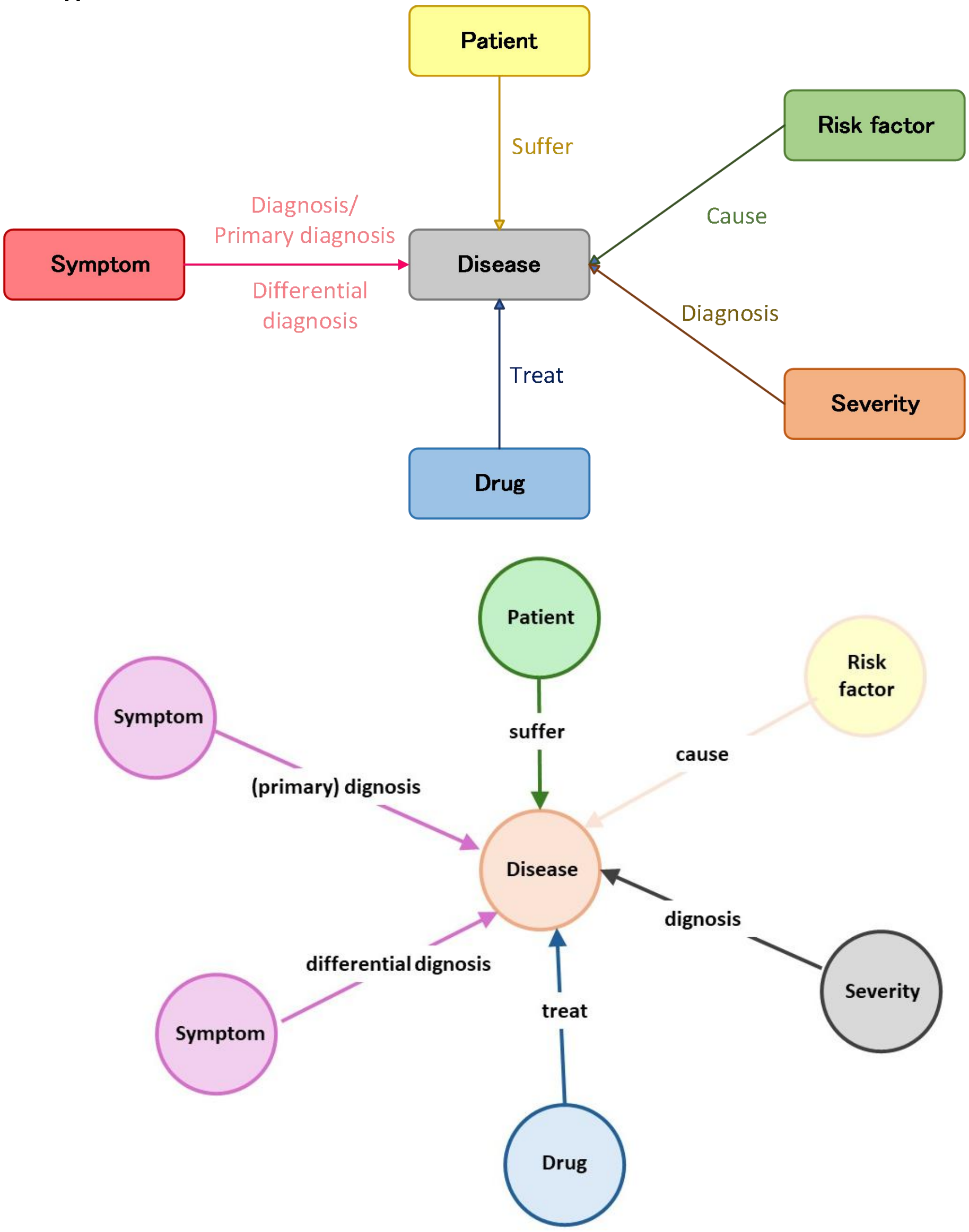

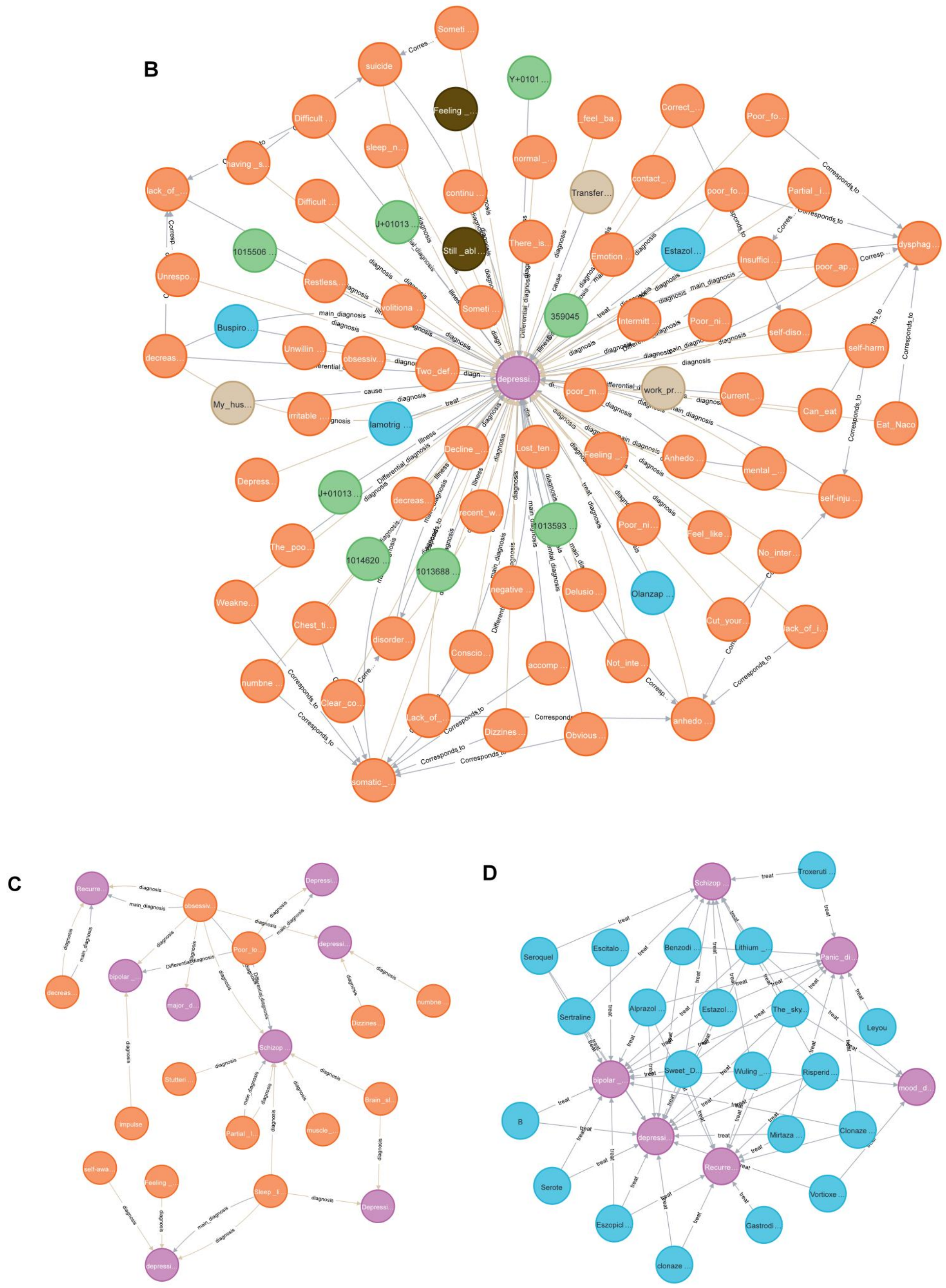

**Figure 3. Construction of KG. (A)** Rectangles represent entity categories, there are six entity categories in total. Arrows and the words next to them represent the relationships between entities, for example, an arrow from the patient category to the disease category represents the "patient suffers disease" triplet. There are a total of seven types of relationships. **(B)** Display of some entities and relationships in the "depression" disease category. Dots represent entities, different entity types are indicated by different colors, and the names of the entities are written on the dots. Arrows represent relationships. **(C, D)** Examples of partial KG structural network. Figure C shows a partial network of nodes related to symptom-disease pairs, and Figure D shows a partial network of nodes related to drug-disease pairs. The complex relationships within the KG form a vast network structure of psychosomatic disorder knowledge.

**Table 2.** Some triplet examples.

| Entity1 | Relationship | Entity2 |
|---|---|---|
| sertraline | treat | depression |
| alprazolam | | |
| low mood | diagnosis | |
| anxiety | | |
| poor relationship with teachers and peers | cause | |
| high work pressure | | |

### *3.2. Nodes Representing Symptom Clusters for Diagnosing Diseases form Modules within the Network Structure.*

We rely on our psychosomatic disorder KG to extract all symptom nodes and their diagnostically related diseases (referred to as “diseases” in this section). We focused on 54 diseases, with a total of 107 types of symptom cluster nodes. We found that for 46 of these 54 diseases, the connectivity components formed by their related symptom clusters were significantly larger than random expectation ($z > 1.5$). See Figure 4.A and Figure 4.E. This indicates that symptoms related to diseases cluster into a local module. Additionally, we analyzed the network separation metric between diseases ($S_{ab}$), where disease pairs overlapping between modules have $S_{ab} < 0$, and those topologically separated between modules have $S_{ab} > 0$. We found the average network separation metric $\langle S_{ab} \rangle$ between diseases and their related symptom modules to be 0.99. See Figure 2.B. The value greater than 0, indicating that symptom modules related to different diseases are distantly separated from each other.

We also explored whether the network distances between symptom modules corresponding to these diseases could reveal clinical relationships between the diseases. For this purpose, we calculated the network distances ($D_{ab}$) between diseases, centered around diseases, with symptom clusters forming modules. We used 5027 symptom-disease pairs to calculate the Co-symptoms count between disease pairs. We found that the Co-symptoms count between two diseases was negatively correlated with the network distance $D_{ab}$ of their symptom modules (Pearson’s correlation = -0.39, $P = 2.26\times10^{-95}$), indicating that a closer network distance between diseases can predict their clinical manifestations to be more similar. See Figure 4.C and Figure 4.E. We also studied whether the network distance between diseases could predict their similarity in psychiatry. For this, we defined semantic similarity of symptoms [24]. We found that the overall semantic similarity of disease pairs negatively correlates with their average network distance $D_{ab}$ (Pearson’s correlation = -0.80, $P < 1\times10^{-100}$). See Figure 4.D. In summary, we found that two diseases with closer network distances share more symptoms and have stronger similarities. The visualization of the symptom modules of diseases is shown in Figure 4.F.

For example, the network distance $D_{ab}$=1.25 between “Mood disorders” and “recurrent depressive disorder” is substantially lower than the average network distance $\langle D_{ab} \rangle$=2.04 for diseases, with 99 Co-symptom count, greatly surpassing the average Co-symptom count across diseases. “Mood disorders” and “recurrent depressive disorder” share many symptoms, such as “irritability” and “sleep disturbances”. Other pairs of diseases with high similarity (highlighted in red in Figure 4.D) include “compulsive disorder” and “depressive episode” ($D_{ab}$=1.67, Co-symptom count=85), among others. Conversely, pairs of diseases with a higher network distance exhibit less Co-symptom count and are not considered similar within the field of psychiatry, such as “acute stress psychosis” and “Tourette's symptoms” (highlighted in green in Figure 4.D), which have a larger $D_{ab}$=3.73 and a lower comorbidity count of 3.

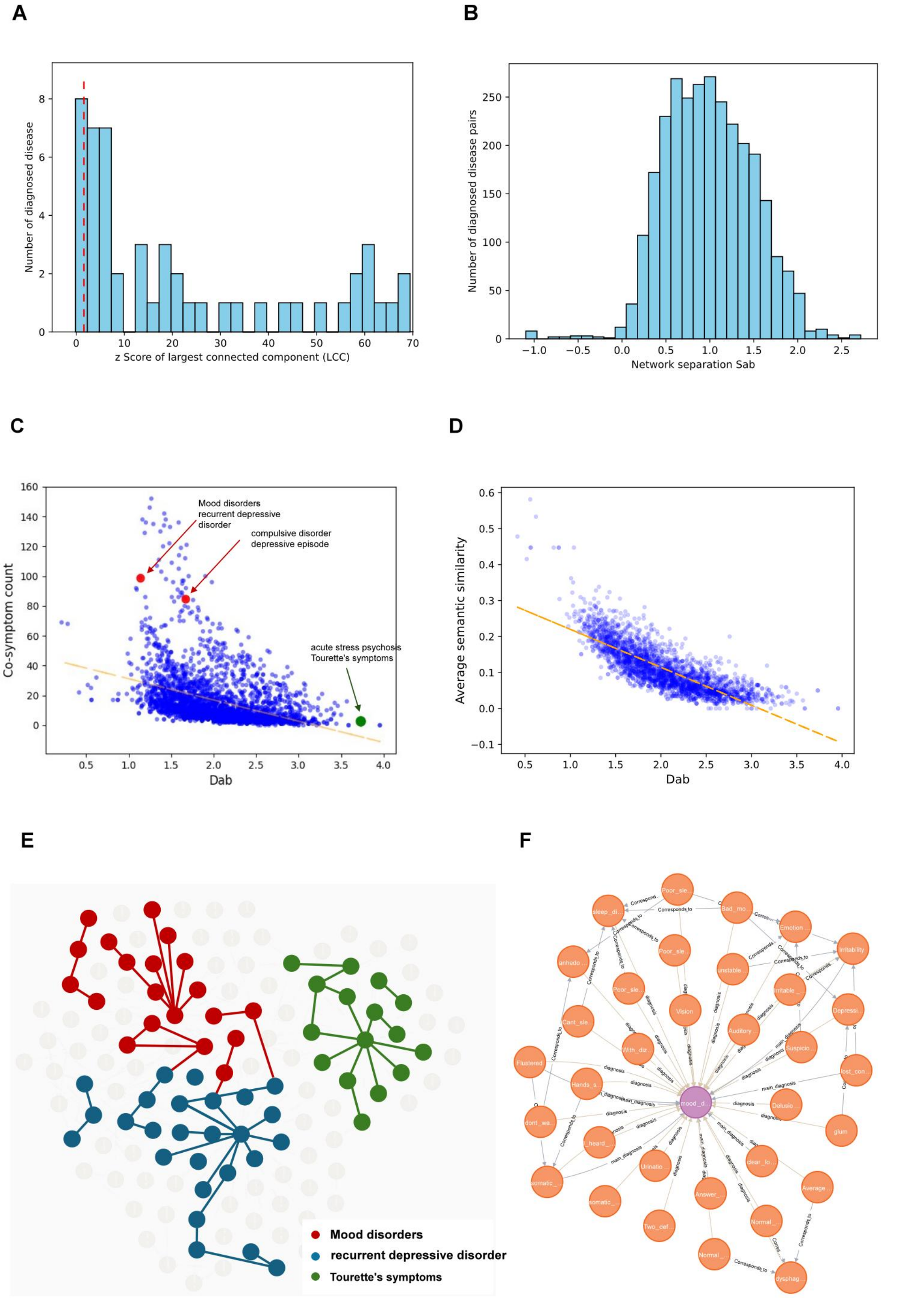

**Figure 4. Symptom module centered on diagnosed diseases. (A)** Distribution of the LCC z-score of the largest connected component formed by the symptoms of 54 diseases. Symptoms of 46 out of the 54 diseases form significantly clustered local modules (z > 1.6). The red dashed line represents z=1.6. **(B)** Distribution of network separation ($S_{ab}$) for symptom groups possessed by all disease pairs, with an average network separation $<S_{ab}> > 0$, indicating that different diseases form symptom modules that are distant from each other. **(C)** The network distance ($D_{ab}$) of interaction between disease pairs and the clinical similarity of diseases (Co-symptom count) are negatively correlated, with Pearson's correlation -0.39. Each dot represents a disease pair. Examples of similar diseases, such as "Mood disorders" and "recurrent depressive disorder" ($D_{ab}$=1.25, Co-symptom count=99), "compulsive disorder" and "depressive episode" ($D_{ab}$=1.67, Co-symptom count=85) are highlighted in red. We also highlight in green an example with a farther network distance and fewer shared symptoms, namely "acute stress disorder" and "Tourette's symptoms" ($D_{ab}$=3.73, Co-symptom count=3). **(D)** The interaction network distance ($D_{ab}$) between disease pairs is negatively correlated with the semantic similarity of symptoms. **(E)** Examples of disease modules and the network distance between disease pairs. Taking "Mood

disorders”, “recurrent depressive disorder”, and “Tourette’s symptoms” as examples. **(F)** Visual representation of the symptom module for the disease "Mood disorders".

### *3.3. Nodes of Disease Groups Diagnosed through Symptoms Cluster into Modules in the Network Structure*

Following the previous section, we again extracted all nodes of symptoms and their diagnosed diseases (referred to as “diseases” in this section). However, we will focus on symptoms as central nodes to observe the local modules formed by disease group nodes. We focused on 105 symptoms, among which there were 54 types of disease group nodes. We found that the connectivity components formed by their related disease groups were significantly larger than random expectation ($z>1.03$). See Figure 5.A, indicating that diseases related to symptoms can also form local modules, meaning there is a many-to-many correspondence between diseases and symptoms. Furthermore, we also found that the $<S_{ab}>$ between the symptom pairs is 0.09, a value close to 0, indicating a high degree of similarity between some symptoms of disease, with some symptom clusters able to reflect a single disease simultaneously. See Figure 5.B

We also explored whether the network distances between disease modules corresponding to these symptoms could reveal the co-occurrence relationships between symptoms. For this purpose, we calculated the $D_{ab}$ between symptoms, centered on the symptoms, with disease groups forming modules. We calculated the number of concurrent diseases between symptom pairs. We found a negative correlation between Co-disease count for two symptoms and their disease modules' network distance $D_{ab}$ (Pearson’s correlation = -0.63, $P < 1\times10^{-100}$). See Figure 5.C. Subsequently, to further validate this co-occurrence relationship, we calculated the relative risk (RR) between each pair of symptoms. RR is a standard measure of the strength of association (in this study, the simultaneous occurrence of two symptoms). We then found a negative correlation between the RR of symptom pairs and their network distance $D_{ab}$ (Pearson’s correlation = -0.29, $P=2.1\times10^{-57}$). See Figure 5.D. This indicates that a closer network distance between symptoms suggests they are more likely to occur together. For example, “decreased interest” and “Emotional discomfort” ($D_{ab}$=0.36, RR=10.7), “decreased interest” and “Vomit” ($D_{ab}$=0.72, RR=9.7) (highlighted in red in Figure 5.D). Conversely, symptom pairs with longer network distances have lower RR and are less likely to co-occur, such as “hypochondriacal delusions” and “Emotional discomfort” ($D_{ab}$=1.83, RR=1.9) (highlighted in green in Figure 5.D). These results validate our hypothesis that the network distances of disease modules associated with symptoms can reflect the co-occurrence relationships between symptoms. The visualization of the disease modules associated with symptoms is shown in Figure 5.E and Figure 5.F.

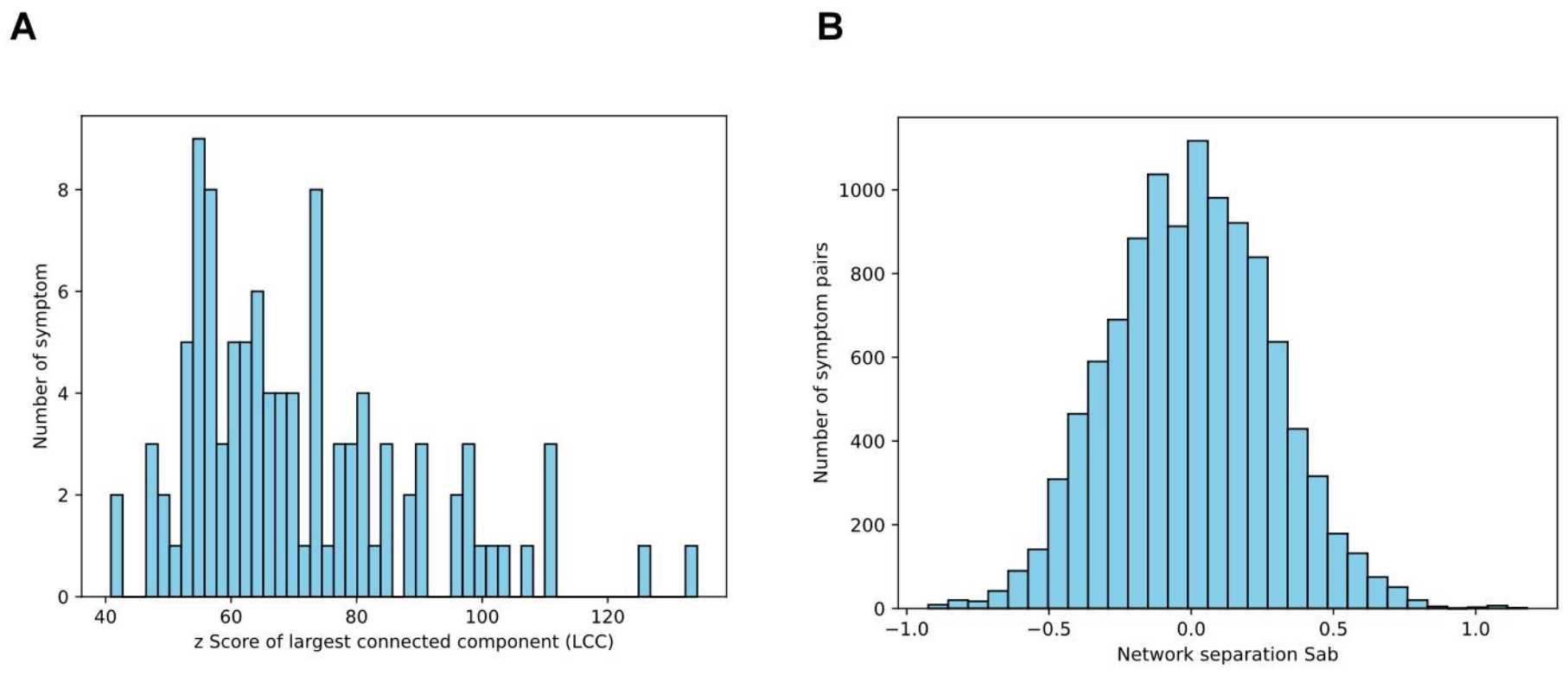

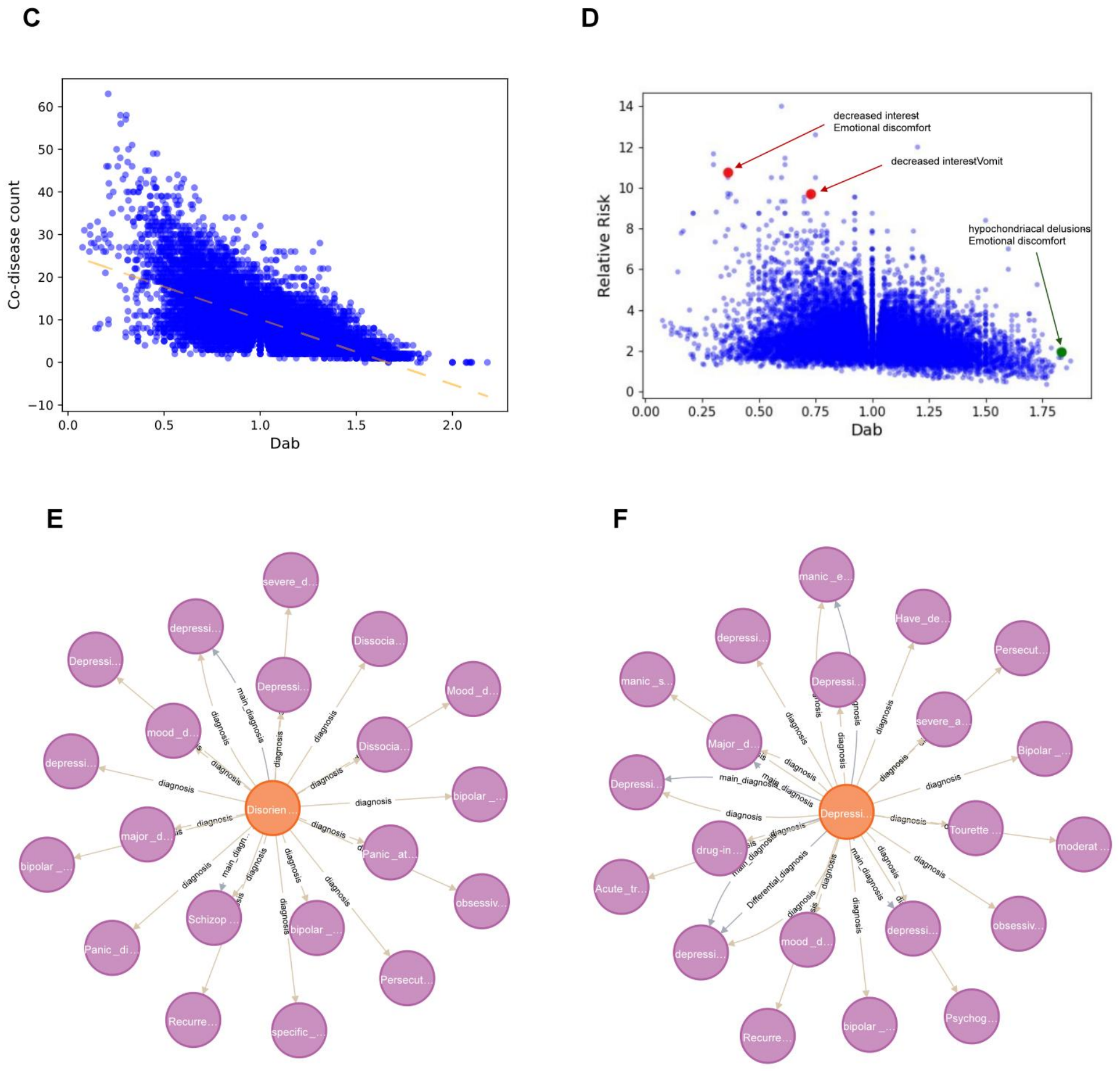

**Figure 5. Symptom-centered diagnostic disease modules. (A)** Distribution of LCC z-score for the largest connected components formed by related diseases of 105 symptoms. **(B)** Distribution of network separation ($S_{ab}$) for all symptom pairs, with an average network separation $\langle S_{ab} \rangle$ very close to 0, indicating a high degree of similarity between some symptoms of psychosomatic disorders. **(C)** The network distance ($D_{ab}$) between symptom pairs and the co-occurrence (Co-disease count) between symptom pairs are negatively correlated (Pearson's correlation = -0.63). **(D)** There is a negative correlation between the relative risk (RR) of symptom pairs and the network distance $D_{ab}$ (Pearson's correlation = -0.29), confirming that shorter network distances between symptoms can predict their co-occurrence. **(E, F)** Visualization of the disease modules associated with the symptoms "Disorientation" and "Depression".

### *3.4. Nodes Representing Groups of Drugs for Diagnosed Diseases form Modules within the Network Structure*

We rely on our psychosomatic disorder KG to extract all drug nodes and their diseases with a therapeutic relationship (referred to as "diseases" in this section). We focused on 37 diseases, among which there were 98 types of drug group nodes. We found that for 35 of these 37 diseases, the connectivity components formed by their related drug clusters were significantly larger than random expectation (z>2.9). See Figure 6.A. This indicates that drugs related to disease treatment cluster into a local module. Furthermore, we also found that the $\langle S_{ab} \rangle$ between diseases and drug modules treating diseases was 1.15, a value greater than 0, indicating that drug modules related to different diseases are distantly separated from each other. See Figure 6.B

We also explored whether the network distances between drug modules corresponding to these diseases could reveal the treatment principles between diseases. For this purpose, we calculated the $D_{ab}$ between diseases, centered on the diseases, with drug groups forming modules. We used 279 drug-disease pairs to calculate the number of drugs shared between pairs of diseases. We found a negative correlation between the Co-drug count for two diseases and their drug modules' network distance $D_{ab}$ (Pearson's

correlation = -0.34, P = 2.08×$10^{-18}$), indicating that a closer network distance between diseases can predict their treatment plans to be more similar. See Figure 6.C. We also studied whether the network distances between diseases could predict their similarity in psychological mechanisms. For this, we defined the semantic similarity of drugs. We found that the overall semantic similarity of disease pairs negatively correlates with their average network distance $D_{ab}$ (Pearson's correlation = -0.82, P < 1×$10^{-100}$). See Figure 6.D. In summary, we found that two diseases with closer network distances have more similar treatment plans and psychological mechanisms. The visualization of the drug modules associated with diseases is shown in Figure 6.E and Figure 6.F.

For example, the network distance $D_{ab}$ between "bipolar disorder" and "depressive episode" is 0.89, much lower than the average network distance of diseases $\langle D_{ab} \rangle$=2.14, with 19 shared drugs, far above the average number of shared drugs for diseases, which is 4. "bipolar disorder" and "depressive episode" share many drugs, such as "seroquel" and "olanzapine". Other disease pairs with high similarity (highlighted in red in Figure 6.C) include "schizophrenia" and "depressive episode" ($D_{ab}$=1.19, co-drugs=15), among others. Conversely, disease pairs with higher network distances have fewer shared drugs and are not considered similar in psychological mechanisms, such as "recurrent depressive disorder" and "Mood disorders" (highlighted in green in Figure 6.C), which have a larger $D_{ab}$=2.04 and a smaller co-drug count of 2.

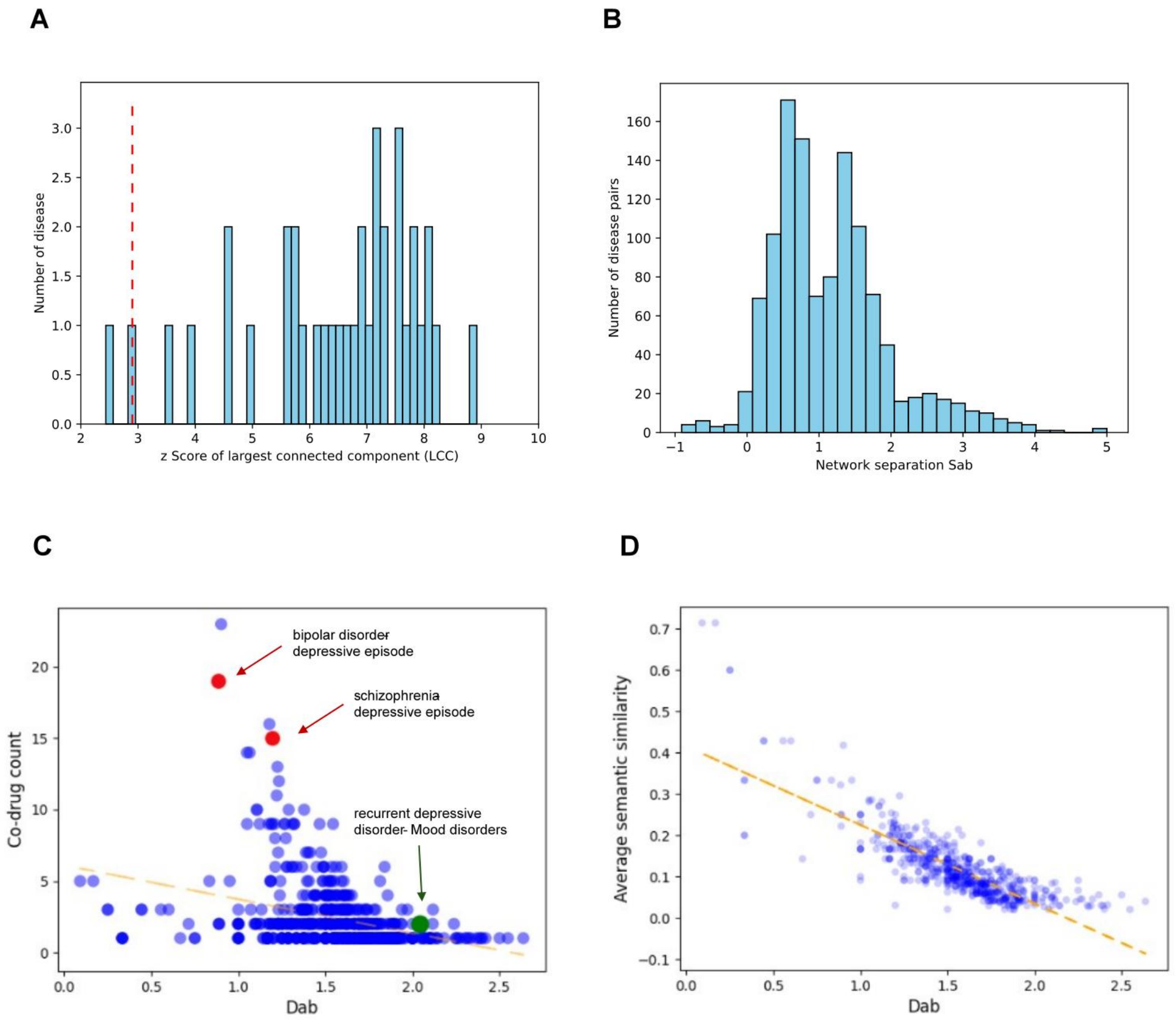

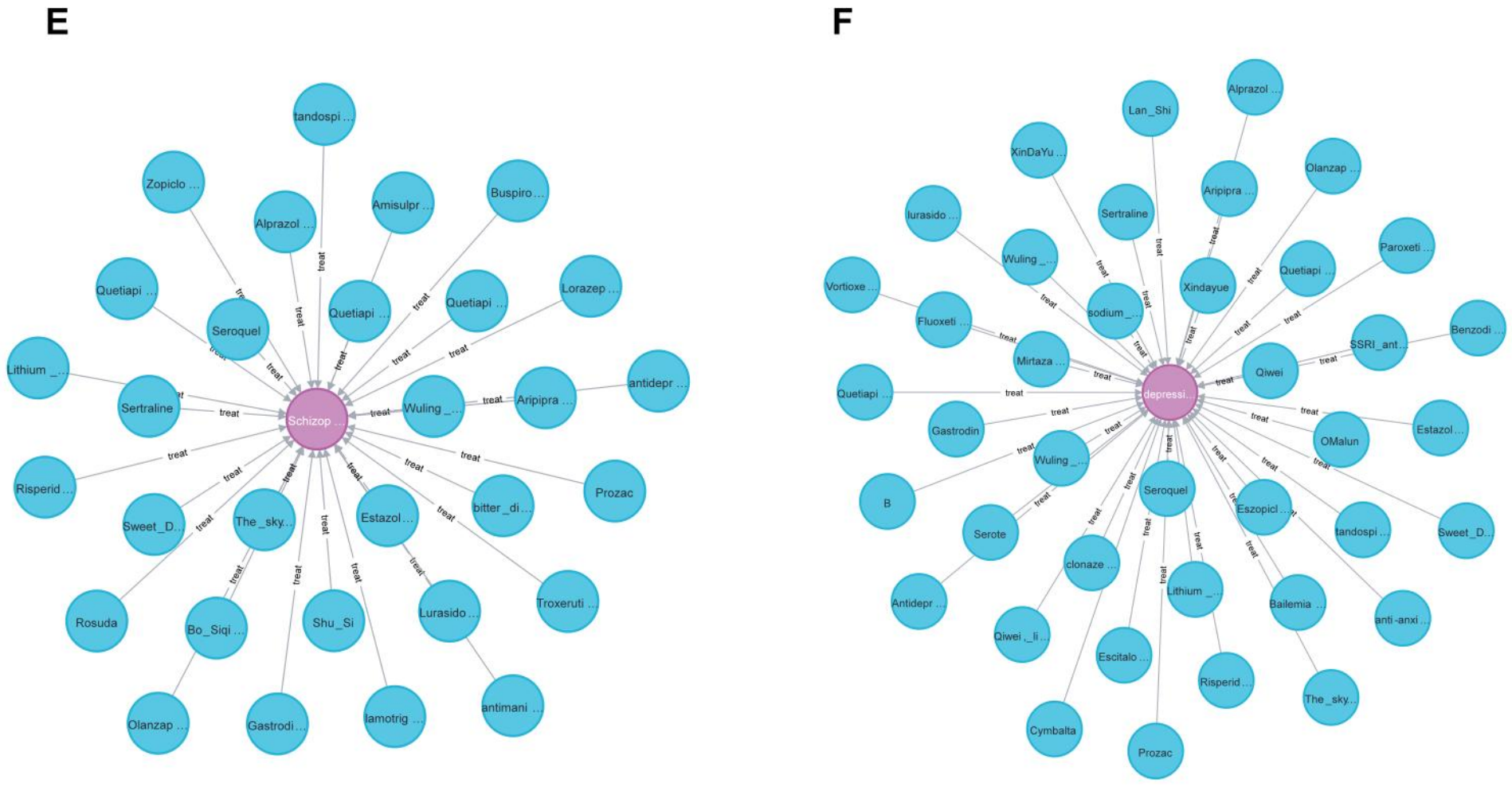

**Figure 6. Drug modules centered on treating diseases. (A)** Distribution of LCC z-score for the largest connected component of treatment-related drugs for 37 diseases. Drugs for 35 out of these 37 diseases form significantly clustered local modules (z > 2.9). The red dashed line represents z=2.9. **(B)** Distribution of network separation ($S_{ab}$) for drug groups corresponding to all disease pairs, with an average network separation $<S_{ab}> > 0$, indicating that different diseases form modules that are distant from each other. **(C)** The network distance ($D_{ab}$) between disease pairs and the similarity of treatment principles (Co-drug count) are negatively correlated, with a Pearson's correlation of -0.34. Each point represents a disease pair. Examples of similar diseases highlighted in red, such as "bipolar disorder" and "depressive episode" ($D_{ab}$=0.89, Co-drug count=19), "schizophrenia" and "depressive episode" ($D_{ab}$=1.19, Co-drug count=15). We also highlighted in green an example with a farther network distance and fewer shared symptoms, namely recurrent "recurrent depressive disorder" and "Mood disorders" ($D_{ab}$=2.04, Co-drug count=2). **(D)** The network distance between disease pairs and the semantic similarity of drugs are negatively correlated, indicating that diseases with closer network distances have similar psychological mechanisms. **(E, F)** Visualization of the drug modules associated with "schizophrenia" and "depressive episode".

### *3.5. The Modules Formed by Symptom Cluster Nodes for Differential Diagnosis of Diseases and the Degree of Association between Diseases and Symptoms*

Besides diagnosed diseases, we rely on our psychosomatic disorder KG to extract all symptoms and their disease nodes with a differential diagnosis relationship. Differential diagnosis means excluding the possibility of a disease based on certain symptoms. We focused on 53 diseases with differential diagnosis, among which there were 243 types of symptom cluster nodes. We found that for 36 of these 53 diseases with differential diagnosis, the connectivity components formed by their related symptom clusters were significantly larger than random expectation (z＞1.1). See Figure 7.A. This indicates that symptoms of diseases with differential diagnosis cluster into a local module. Furthermore, we also found the $<S_{ab}>$ between symptom modules corresponding to diseases with differential diagnosis to be 2.26, a value greater than 0, indicating that symptom modules corresponding to different differential diagnosis diseases are far apart from each other. See Figure 7.B

In our psychosomatic disorder KG, diagnostic relationships are divided into primary diagnosis relationships and diagnosis relationships. The symptoms connected by diagnosis relationships are mostly patient-reported and colloquial, whereas those connected by primary diagnosis relationships are summarized by psychiatrists and are terms standardized in the field of psychiatric medicine. To explore the degree of association between symptoms and their diseases for both types of relationships, we observed the network structure of symptoms and diseases under primary diagnosis relationships and diagnosis relationships, respectively. To quantify the degree of association, we used two metrics [18]: (1) proximity distance d, the average shortest distance between nodes within symptom and disease modules. (2) the proximity z score, measuring the difference between proximity distance d and random expectation. For these metrics, lower values indicate a higher degree of association between diseases and symptoms. For the proximity z score, z < 0 indicates closer than random, z > 0 indicates further than random. We plotted box plots for both types of relationships, comparing symptom-disease pairs in primary diagnosis relationships (orange bars) and diagnosis relationships (blue bars). See Figure 7.C. We found that under both metrics, the orange bars are consistently lower than the blue bars, indicating that symptom-disease pairs in primary diagnosis relationships have a stronger degree of association and are of higher reference value. Specific examples are shown in Figure 7.D.

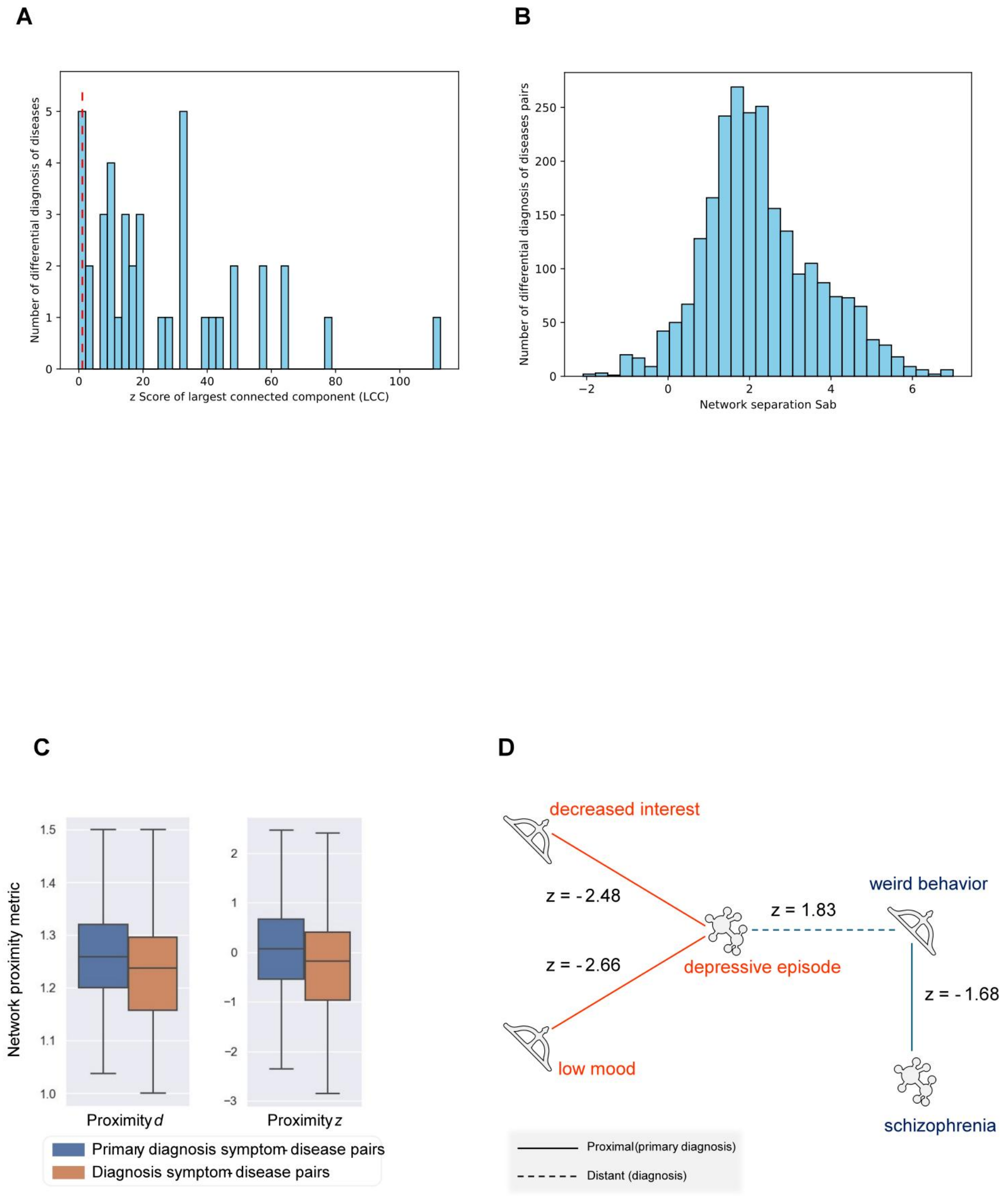

**Figure 7. Analysis of primary diagnosis, diagnosis, and differential diagnosis. (A)** Distribution of LCC z-score for the largest connected components formed by symptoms corresponding to 53 differential diagnosis diseases. Of these, symptoms of 36 diseases form significantly clustered local modules (z>1.1). The red dashed line represents z=1.1. **(B)** Distribution of network separation ($S_{ab}$) for symptom clusters corresponding to all differential diagnosis disease pairs, with an average network separation $\langle S_{ab} \rangle$ greater than 0, indicating that different differential diagnosis diseases form modules distant from each other. **(C)** Disease-symptom pairs are divided into primary diagnosis relationships and diagnosis relationships. Symptom-disease pairs in the primary diagnosis relationship (orange bars) show a shorter network distance than those in the diagnosis relationship (blue bars), indicating a stronger association in primary diagnosis relationships. **(D)** For example, symptoms such as "decreased interest" and "low mood" primarily diagnose "depressive episode" (with a very low network proximity z score). They are included in the main

clinical manifestations of "depressive episode" in real life. However, the symptom of "weird behavior", having a higher network proximity z score, indicates that it is not a main symptom of "depressive episode" but a main symptom of "schizophrenia".

### *3.6 Comparison of Results from Different Models*

We evaluate the NER task performance on the RoBERTa, LLaMA, and LLaMA fine-tuned with LoRA. We measured the NER task quality in terms of precision (P), recall (R), and the F1 score. Table 3 shows the evaluation result of different models in the NER task. It indicates that the LLaMA fine-tuned with LoRA has a clear advantage in the NER task, showing that this method can more effectively identify entities in unstructured text.

**Table 3.** Evaluation results of different models in NER task

| Model | Precision | Recall | F1 |
|---|---|---|---|
| BERT | 0.91 | 0.63 | 0.74 |
| LLM | 0.86 | 0.71 | 0.78 |
| LLM fine-tuned by LoRA | 0.93 | 0.76 | 0.83 |

We evaluate the Q&A task performance of LoRA on Chatglm2 and LLaMA, and Freeze on Chatglm2 and LLaMA. We measured the Q&A task quality in terms of BLEU, ROUGE-1, ROUGE-2, ROUGE-L, Samples per Second and Steps per Second. The first four metrics are used to evaluate the quality of answers in the Q&A task. The latter two metrics are used to evaluate the efficiency and performance of the model.

**Table 4.** Evaluation results of different models and methods in Q&A task

| Model & Method | BLEU | ROUGE-1 | ROUGE-2 | ROUGE-L | Samples per Second | Steps per Second |
|---|---|---|---|---|---|---|
| Chatglm2 | 4.650 | 19.175 | 5.417 | 11.31 | 0.241 | 0.127 |
| Chatglm2 (LoRA) | 80.579 | 87.8248 | 80.3199 | 85.81 | 0.338 | 0.169 |
| Chatglm2 (Freeze) | 25.428 | 30.1160 | 3.7143 | 40.39 | 1.995 | 0.344 |
| LLaMA | 2.993 | 14.630 | 4.399 | 9.399 | 0.127 | 0.064 |
| LLaMA (LoRA) | 78.182 | 83.810 | 78.330 | 83.965 | 1.868 | 0.234 |
| LLaMA(Freeze) | 50.993 | 55.629 | 50.339 | 52.392 | 1.26 | 0.123 |

Table 4 shows the evaluation result of different models and methods in the NER task. We can see that the two baselines fine-tuned with LoRA perform excellently in Q&A quality. Moreover, Chatglm2 outperforms LLaMA by a large margin, indicating that Chatglm2 is more suitable for the third dataset. Therefore, we use the fine-tuned Chatglm2 as the LLM for the knowledge base.

In the Knowledge Base Q&A task, we use three text embedding models: m3e, text2vec-large-chinese, and bge-large-zh. By using and comparing the results of these three text embedding models, we determine the suitable type of text embedding for this study, thereby optimizing the knowledge base system. We measured the Knowledge Base Q&A task quality in terms of ROUGE-Rec, ROUGE-Pre, ROUGE-1F1, and BLEU-4. The results are shown in Table 5. It shows that m3e-base is the best text embedding model, achieving high scores across all metrics.

**Table 5.** Evaluation results of different text embedding models Knowledge Base Q&A task

| Model | ROUGE-Rec | ROUGE-Pre | ROUGE-1F1 | BLEU-4. |
|---|---|---|---|---|
| m3e | 0.8293 | 0.4546 | 0.63 | 0.530 |
| text2vec-large-chinese | 0.246 | 0.428 | 0.218 | 0.0004 |
| bge-large-zh | 0.326 | 0.3222 | 0.324 | 0.066 |

## 4. Discussion

In this study, we defined entity and relation types based on clinical terminology widely used in the field of psychiatry. Through communication and discussion, we developed a comprehensive set of standardized concepts that can precisely represent various clinical entities and their relationships. To accommodate the demands of real-some mod world psychiatric applications, we also sought opinions from domain experts and made ifications and adjustments to the entity and relation types. Ultimately, we obtained 6 types of entities and 7 types of relationships.

Due to information repetition, the extracted knowledge on psychosomatic disorders inevitably contains a certain degree of redundancy. Therefore, we used entity alignment methods to remove duplicate entities and standardized and unified the entities. As a result, we constructed a high-quality psychosomatic disorder KG. After constructing the KG, we analyzed its structure, revealing the connections between diseases, symptoms, and drugs in the field of psychosomatic disorders.

We found that in 54 diseases, the symptoms corresponding to 46 diseases, diseases related to 105 symptoms, and the drugs required for 35 of 37 diseases all formed significantly clustered local modules. This phenomenon indicates that there is a close interrelation between symptoms and pharmacological treatments in the diagnosis and treatment of psychosomatic disorders, suggesting that the network structure of psychosomatic disorders might have a decisive influence on symptom and drug selection. The discovery of this correlation not only unveils the intrinsic link between the symptoms and treatment of psychosomatic disorders but also provides a theoretical basis for using bioinformatics methods to predict potential treatment strategies for mental diseases. For example, by analyzing the symptom cluster modules, we can identify which symptoms are key manifestations of a particular psychosomatic disorder and thereby infer the most effective medication combinations.

The symptom modules of diseases reflect their clinical similarity. For instance, the diseases "Mood disorders" and "recurrent depressive disorder" have close network distances $D_{ab}$ in their symptom modules and a high number of shared symptoms, hence a high clinical similarity between them. Conversely, "acute stress disorder" and "Tourette syndrome" have a high network distance $D_{ab}$ and very few shared symptoms, therefore are not considered similar by the field of psychiatry. By analyzing the clinical similarities between diseases, it aids doctors in more accurately distinguishing and identifying similar diseases during the early diagnostic stages, especially in cases of overlapping or unclear symptoms, thus improving diagnostic accuracy. Moreover, it provides a symptom-based method of disease classification that may complement traditional etiological classification methods, offering data support for new classification standards in psychiatric disorders.

The disease modules of symptoms reflect the co-occurrence relationships among symptoms. For example, the symptoms "hallucinations" and "anhedonia" have close network distance $D_{ab}$ in their disease modules and a high number of concurrent diseases, indicating a high degree of co-occurrence between the symptoms. Conversely, "elevated mood" and "lack of insight" have a high network distance $D_{ab}$ and very few concurrent diseases, thus exhibiting a low degree of co-occurrence. Subsequently, we validated this conclusion using the relative risk (RR) of symptoms, showing that the co-occurrence of symptoms is negatively correlated with their network distance between modules. By analyzing the co-occurrence relationships of symptoms between different diseases, it helps to understand why certain symptoms frequently appear together, potentially indicating a common pathophysiological basis. Furthermore, in the management of patients with multiple psychosomatic disorders, it enables a better understanding and prediction of symptom progression, optimizing treatment plans.

The drug modules of diseases reflect the similarity in treatment approaches among diseases. For example, the diseases "bipolar disorder" and "depressive episode" have close network distance $D_{ab}$ in their drug modules and share many drugs, indicating a high similarity in their treatment approaches. Conversely, "recurrent depressive disorder" and "Mood disorders" have a high network distance $D_{ab}$ in their drug modules and share very few drugs, thus their treatment plans are generally not comparable. By analyzing the similarity in treatment plans between different diseases, it is possible to discover potential new medication guidelines or alternative treatment methods. Furthermore, this can assist medical researchers in extending the known effects of drugs to new disease areas, promoting drug repurposing research.

For diagnosed disease-symptom pairs and disease-drug pairs, we have also defined semantic similarity to analyze the degree of similarity between diseases. This definition allows for the quantification of the associations between diseases and symptoms, and diseases and drugs. Using this method helps doctors more precisely identify and differentiate various psychosomatic disorders, which is crucial for improving diagnostic accuracy, especially among diseases with similar manifestations such as various anxiety or Mood disorders. Furthermore, it can help uncover potential similarities in drug responses among different diseases, thereby offering personalized treatment options, particularly valuable for psychosomatic disorders that require trials of multiple medications to find an effective treatment plan.

For primary and diagnostic disease-symptom pairs, our analysis using proximity distance d and proximity z score found that primary diagnosis relationships have stronger associations, confirming that network proximity can effectively predict disease-symptom pairs with stronger correlations. For example, the symptoms "reduced interest" and "lack of will" in the primary diagnosed disease "depressive episode" have a low proximity distance d and proximity z score, and these are major clinical manifestations of "depressive episode" in real life. Meanwhile, the symptom "bizarre behavior" and the disease "depressive episode" have a high proximity distance d and z score, but it has a low proximity distance d and z score with "schizophrenia", indicating it is not a main symptom of depressive episode but of schizophrenia. This analysis helps medical professionals more precisely identify the core symptoms associated with specific psychosomatic disorders. This is crucial for the diagnosis of psychosomatic disorders, especially in the early stages, as accurate symptom identification can significantly improve the success rate of treatment.

Our work lays the foundation for further development of smart medical information systems in psychiatry. medical information intelligent systems greatly lack specially constructed domain knowledge bases as accurate sources of knowledge. Unlike general encyclopedic knowledge, our knowledge originates from the summarized medical records of psychiatrists, constructing a highly specialized knowledge graph database. The KG we constructed helps guide psychiatric staff to engage more effectively with patients, analyze the proximity of knowledge modules in the dialogues formed, and enhance the quality of healthcare. Furthermore, by analyzing the knowledge network structure of psychosomatic disorder, we have delved into the connections between these diseases, symptoms, and drugs, providing a technical roadmap and foundational data for developing applications that save psychiatrists' time, enhance treatment efficacy and compliance, and improve patient quality of life.

## 5. Conclusion

This research collected 262 cases from the Psychiatric Department of Zhongda Hospital in Nanjing, China and used BERT and LLM for entity extraction to build a psychosomatic disorder KG containing 3285 entities and 9668 relationships. Subsequently, graph theory was applied to analyze the structure of the constructed KG. The study found that symptoms of diseases, diseases related to these symptoms, and the drugs required for these diseases form local clustering modules within the graph structure. Semantic similarity analysis was also defined to measure the degree of similarity between diseases. Through this definition, the associations between diseases and symptoms, and diseases and drugs can be quantified. The research findings are as follows:

(1) The LLM, fine-tuned with LORA, improved in its ability to extract entities, achieving an accuracy close to that of the BERT model, up to 93%.

(2) The average network separation $S_{ab}$ measure between disease and related symptom modules is 0.99, greater than 0, indicating that symptom modules associated with different diseases are far apart from each other. The Co-symptom count between two diseases and the network distance $D_{ab}$ of their symptom modules are negatively correlated. The overall semantic similarity between disease pairs and their average network distance $D_{ab}$ is also negatively correlated, suggesting that closer network distances between diseases can predict more similar clinical presentations.

(3) The average network separation $S_{ab}$ measure between symptoms and their related disease modules is 0.09, close to 0, indicating that a high degree of similarity among some symptoms can reflect the same disease. The RR between pairs of symptoms is negatively correlated with their network distance $D_{ab}$, suggesting that symptoms are more likely to co-occur with closer network distances.

(4) The average network separation $S_{ab}$ measure between diseases and their drug treatment modules is 1.15, greater than 0, indicating that drug modules associated with different diseases are significantly distant from each other. The Co-drug count between diseases and the network distance $D_{ab}$ of their drug modules are negatively correlated. The overall semantic similarity between disease pairs and their average network distance $D_{ab}$ is also negatively correlated, suggesting that diseases with closer network distances have more similar treatment regimens and psychological mechanisms.

(5) By comparing the proximity d and proximity z score metrics, it is shown that symptom-disease pairs in primary diagnostic relationships have a stronger association and higher referential value than those in diagnostic relationships.

The research results help medical professionals more accurately identify the core symptoms of diseases, not only revealing the interrelationships among psychosomatic disorders but also potentially providing a theoretical basis for developing new treatment methods and improving existing treatment strategies.

**Author Contributions:** Conceptualization, Zihan Zhou, Ziyi Zeng and Yihui Zhu; methodology, Zihan Zhou, Ziyi Zeng and Yihui Zhu; formal analysis, Ziyi Zeng; data curation, Ziyi Zeng, Wenhao Jiang, Yihui Zhu, Jiaxin Mao, Yonggui Yuan, Zhiyuan Ren, Tongtong Xu, Xiangxiang Pei, Mengyu Yao and Yunqian Chen; writing—original draft preparation, Zihan Zhou, Ziyi Zeng; writing—review and editing, Zihan Zhou, Ziyi Zeng, Yihui Zhu, Wenhao Jiang, Jiaxin Mao, Yonggui Yuan, Min Xia and Shubin Zhao; visualization,Zihan Zhou, Ziyi Zeng and Yihui Zhu; supervision, Min Xia and Shubin Zhao. All authors have read and agreed to the published version of the manuscript.

**Funding:** This work was supported by the National Natural Science Foundation of China, China (Grant No.12202210) and the Startup Foundation for Introducing Talent of Nanjing University of Information Science and Technology, China (Grant No.2022r095).

**Institutional Review Board Statement:** Ethics Committee Approval given by the Clinical Research Ethics Committee, Zhongda Hospital Affiliated to Southeast University, Nanjing in accordance with the principles of the Helsinki Declaration of 2013 (protocol code 2021ZDSYLL349-P02, 29 April 2022).

**Informed Consent Statement:** Patient consent was waived because the study was conducted on past collected medical records. All records were anonymous and can not be connect with existing identification information.

**Data Availability Statement:** The data that support the findings of this study are available from the corresponding author (Yihui Zhu) upon request. The data are not publicly available due to [ethical restriction].

**Acknowledgments:** The author is very grateful for the guidance provided by his mentor Yihui Zhu, the encouragement given by his relatives, and the contributions of Tantao Su and Ruoliu Xu, Yihan Du and Shijie Peng to this research.

**Conflicts of Interest:** The authors declare no conflict of interest.